\pdfoutput=1

\documentclass[twocolumn]{article}         %
\usepackage[total={7.2in, 9in}]{geometry}
\usepackage{graphicx}
\usepackage{amsopn}
\usepackage{amsmath,amssymb} 
\usepackage{amsfonts}
\usepackage{comment}
\usepackage{xcolor,colortbl}
\usepackage{multirow}
\usepackage{xspace}
\usepackage{mathtools}
\usepackage{algorithm}
\usepackage[noend]{algpseudocode}
\usepackage{afterpage}%
\usepackage{natbib}
\usepackage{hyperref}
\usepackage{tikz}
\algnewcommand{\IfThenElse}[3]{%
  \State \algorithmicif\ #1\ \algorithmicthen\ #2\ \algorithmicelse\ #3}
  \algnewcommand{\DoWhile}[2]{%
    \State \algorithmicdo\ #1\
    \State\algorithmicwhile\ #2}

\newcommand{\trans}{^{\!\mathsf{T}}}

\renewcommand{\div}{\nabla\!\cdot\!}
\newcommand{\RNum}[1]{{\bf (\lowercase\expandafter{\romannumeral #1\relax})}}
\newcommand{\dx}{\mathrm{d}x}

\newcommand{\uf}{u}

\newcommand{\Alg}{Alg.\@\xspace}
\newcommand{\Eq}{Eq.\@\xspace}

\newcommand{\Fig}{Fig.\@\xspace}

\newcommand{\Sec}{Sec.\@\xspace}
\newcommand{\Tab}{Tab.\@\xspace}
\newcommand{\eg}{e.g.\@\xspace}
\newcommand{\Eg}{E.g.\@\xspace}

\newcommand{\ie}{i.e.\@\xspace}

\newcommand{\wrt}{w.r.t.\@\xspace}
\newcommand{\etc}{etc.\@\xspace}
\newcommand{\cf}{c.f.\@\xspace}
\DeclareMathOperator*{\argmin}{arg\,min}

\definecolor{myRed}{rgb}{0.6,0.1,0.}
\definecolor{myBlue}{rgb}{0.0,0.2,0.6}
\definecolor{myGreen}{rgb}{0.2,0.6,0.}

\newcommand{\changes}[1]{\textcolor{black}{#1}\xspace}
\newcommand{\invisible}[1]{}

\newcommand{\tablecaptionspace}{2mm}

\begin{document}

\title{3D Fluid Flow Estimation with Integrated Particle Reconstruction%
}

\author{Katrin Lasinger$^1$, Christoph Vogel$^2$, Thomas Pock$^{2,3}$ and Konrad Schindler$^1$
\vspace{5pt}
\\
\small $^1$Photogrammetry and Remote Sensing, ETH Zurich, Switzerland \\
\small $^2$Institute of Computer Graphics \& Vision, TU Graz, Austria \\
\small $^3$Austrian Institute of Technology, Austria
}
\date{}

\twocolumn[
 \begin{@twocolumnfalse}
 \maketitle
\begin{abstract}
The standard approach to densely reconstruct the motion in a volume of fluid
is to inject high-contrast tracer particles and record their motion
with multiple high-speed cameras.
Almost all existing work processes the acquired multi-view video
in two separate steps, utilizing either a pure Eulerian or pure Lagrangian approach.
Eulerian methods perform a voxel-based reconstruction of particles per time step,
followed by 3D motion estimation, with some form of dense matching
between the precomputed voxel grids from different time steps.
In this sequential procedure, the first step cannot use temporal
consistency considerations to support the reconstruction, while the
second step has no access to the original, high-resolution image data.
Alternatively, Lagrangian methods reconstruct an explicit, sparse set of particles
and track the individual particles over time.
Physical constraints can only be incorporated in a post-processing step when interpolating 
the particle tracks to a dense motion field.
We show, for the first time, how to \emph{jointly} reconstruct both
the individual tracer particles and a dense 3D fluid motion field from
the image data, using an integrated energy minimization.
Our hybrid Lagrangian/Eulerian model reconstructs
individual particles, and at the same time recovers a dense 3D motion
field in the entire domain.
Making particles explicit greatly reduces the memory consumption and
allows one to use the high-resolution input images for
matching. Whereas the dense motion field makes it possible to include
physical a-priori constraints and account for the incompressibility
and viscosity of the fluid.
The method exhibits greatly ($\approx\!70\%$) improved results over our
recently published baseline with two separate steps for 3D reconstruction and
motion estimation. Our results with only two time steps are
comparable to those of state-of-the-art tracking-based methods that
require much longer sequences.

\end{abstract}

 \end{@twocolumnfalse}
 
\vspace{10pt}
]

\section{Introduction}
\label{sec:intro}

Motion estimation of fluids is a challenging task in experimental fluid mechanics with important applications in research and industry: 
Observations of fluid motion and fluid-structure
interaction form the basis of experimental fluid dynamics.
Measuring and understanding flow and turbulence patterns is important
for aero- and hydrodynamics in the automotive, aeronautic and
ship-building industries, \eg to design efficient shapes and
to test the elasticity of components.
Biological sciences are another application field, \eg, behavioral
studies about aquatic organisms that live in flowing water, %
or the
analysis of the flight dynamics of birds and
insects~\citep{mic-15,wu-09}.

\begin{figure*}[t]
 \centering
 \includegraphics[width=\textwidth]{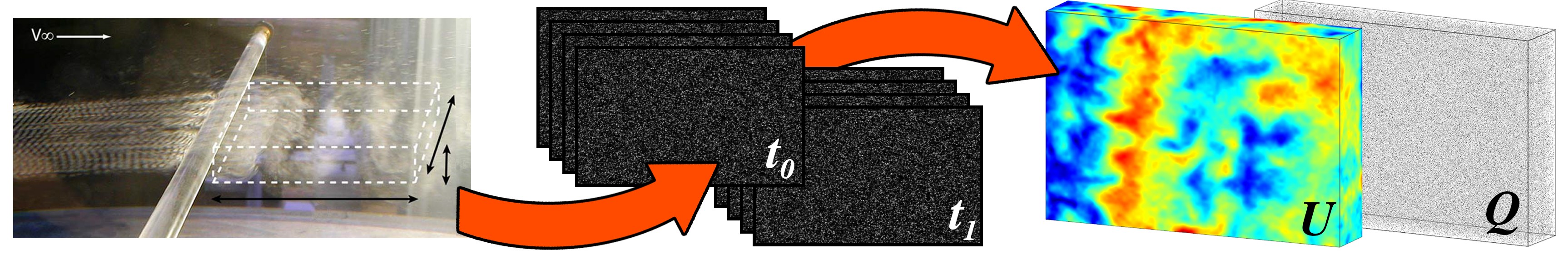}
 \caption{From
  2D images of a fluid injected with tracer particles, recorded at two
  consecutive time steps, we jointly reconstruct the particles
  $\mathcal{Q}$ and a dense 3D motion field $\mathcal{U}$. The example
  shows the experimental setup
  of~\citet{mic-06}.}  \label{fig:overview}
\end{figure*}

To capture fluid motion, one common approach is to inject tracer particles into the fluid,
which densely cover the illuminated area of interest, and to record
them with one or multiple high-speed cameras. %
From this idea two strategies have evolved in experimental fluid mechanics: \emph{particle image velocimetry} (PIV) and \emph{particle tracking velocimetry} (PTV) \citep{adr-11,raf-13}.
PIV outputs a dense velocity field (Eulerian view), while PTV computes a flow vector at each particle location (Lagrangian view).

In the standard 2D-PIV setup, a thin slice of the measurement area is illuminated by a laser scanner, allowing only for 2D in-plane motion estimation. Motion vectors are recovered on a dense regular grid by correspondence matching of local support windows (in PIV terminology ``interrogation windows''). 
In contrast, \emph{particle tracking velocimetry} (PTV) employs a Lagrangian view of the problem by tracking individual particles over multiple time steps and effectively describing the motion of the fluid only on those sparse locations. Since individual particles have to be identified, the maximum allowed seeding density is usually lower than for PIV measurements, where particle constellations are matched instead.

In the 3D setup, a measurement volume is illuminated with a thick laser slice and observed by synchronized cameras from multiple viewpoints.
We illustrate the basic setup in \Fig~\ref{fig:overview}: 3D Particle locations and a dense motion field are recovered from a set of input images from two consecutive time steps.
There is a trade-off regarding the \changes{seeding} density of the particles:
A higher density delivers an increased effective spatial resolution,
but also raises the ambiguity of the matching.

\changes{
Both Eulerian and Lagrangian approches have been proposed to tackle the problem in 3D (\cf \citet{maa-93,els-06,Champagnat2011,dis-12,che-14,las-17,sch-16}), coming with their individual architectural problems. Eulerian approaches perform 3D reconstruction and motion estimation in two sequential steps and require large voxel volumes to represent the high-frequency particle data. Lagrangian approaches, on the other hand, require tracking over multiple time steps to resolve ambiguities in the triangulation, and cannot directly incorporate physical constraints.}

In this work we propose a \emph{joint energy model} for the
reconstruction of the particles and the 3D motion field from two time steps, so as to
capture the inherent mutual dependencies.
The model uses the full information -- all available images from both
time steps at full resolution -- to solve the problem. We opt for a
hybrid \emph{Lagrangian/Eulerian} formulation: particles are modeled
individually, while the motion field is represented on a dense grid.
Recovering explicit particle locations and intensities sidesteps the
costly 3D parameterization of voxel occupancy, as well as the
use of a large matching window. Instead, it directly compares evidence
for single particles in the images, yielding significantly higher accuracy.

To represent the motion field, we opt for a trilinear finite element
basis.
Modeling the 3D motion densely allows us to incorporate
physical priors that account for incompressibility
and viscosity of the observed fluid, similar to our previous work \citep{las-17}.
This can be done efficiently, at a much lower voxel resolution than
would be required for particle reconstruction, due to the smoothness of
the 3D motion field \citep{che-14,las-17}.

We model our problem in a variational setting. To better resolve
particle ambiguities, we add a prior to our energy that encourages
sparsity.
In order to overcome weak minima of the non-convex energy, we include
a proposal generation step that detects putative particles in the
residual images, and alternates with the energy minimization.
For the optimization itself, we can rely on the very
efficient \emph{inertial Proximal Alternating Linearized Minimization}
(iPALM) \citep{Bolte2014,poc-16}. It is guaranteed to converge to a
stationary point of the energy and has a low memory footprint, so that we
can reconstruct large volumes.

Compared to the baselines of \citet{els-06} and our own previous work \citep{las-17}, which both
address the problem sequentially with two independent steps, we
achieve particle reconstructions with higher precision and recall, at
all tested particle densities; as well as greatly improved motion estimates.
The estimated fluid flow visually appears on par with state-of-the-art
techniques like \citet{sch-16}, which use tracking over
multiple time steps and highly engineered
post-processing \citep{schn-16,ges-16}.

The present paper is based on the preliminary version \citep{las-18}.
\changes{
We have extended the original approach by integrating also the polynomial camera model by \citet{sol-97}. This model compensates for refractions at air-glass and glass-water transitions and thus allows for experiments with liquids, such as water. Furthermore, we have added additional experiments and visualizations for setups both in water and air.}

\section{Related Work}
\label{sec:relatedWork}

We focus here on 3D-PIV and PTV approaches, and refer to \citet{adr-11} or \citet{raf-13} for an exhaustive review of 2D techniques.
The first method to operate on 3D fluid volumes
was \changes{the Lagrangian} 3D-PTV \changes{method} \citep{maa-93}, where individual particles are detected in
different views, triangulated and tracked over time.  Yet, as the \changes{seeding} density increases the particles quickly start to overlap in
the images, leading to ambiguities. Therefore, 3D-PTV is only recommended
for densities $<0.005$ ppp (particles per pixel). To handle higher \changes{seeding}
densities, \citet{els-06} introduced \changes{the Eulerian} Tomo-PIV \changes{method}. They first
employ a tomographic reconstruction (\eg MART) \citep{atk-09} per time
step, to obtain a 3D voxel space of occupancy
probabilities. Cross-correlation with large local 3D windows ($\geq
35^3$) \citep{Champagnat2011,dis-12,che-14} then yields the flow.
Effectively, this amounts to matching particle constellations,
assuming constant flow in large windows, which smooths the output to
low effective resolution.
Recently, a new \changes{Lagrangian} particle tracking method \textit{Shake-the-Box} (StB)
was introduced \citep{sch-16}. It builds on the idea of
\textit{iterative particle reconstruction} (IPR) \citep{wie-13}, where
triangulation is performed iteratively, with a local position and
intensity refinement after every triangulation step.
In subsequent time steps, particle locations are predicted from trajectory information from previous time steps, and refined by small perturbations in all directions ("shaking") to find the location with the lowest reprojection error.

None of the above methods accounts for the physics of (incompressible)
fluids during reconstruction. In StB \citep{sch-16}, a post-process
interpolates sparse tracks to a regular grid. At that step, but not
during reconstruction, physical constraints can be included \citep{schn-16,ges-16}.
Variational approaches that impose physically consistent regularizers were first proposed for the 2D PIV setup. %
\changes{
\citet{ruh-07} presented an optical flow-based approach that incorporates physical priors using the Stokes equations.
In \citep{ruh-06} this idea was further extended to the full Navier-Stokes equations. With the help of the vorticity transport equation they obtain a spatio-temporal regularization that can model turbulent fluids. A drawback of the method is that the vorticity is usually not known beforehand, thus it has to be initialized with $\omega = 0$, and observations from several time steps are needed for the estimation to converge ($\approx 5$ in their experiments).
}

\changes{In our earlier work \citep{las-17} we have proposed a 3D variational approach that combines} TomoPIV with variational 3D flow estimation. \changes{We}
account for physical constraints with a regularizer derived from the
stationary Stokes equations, similar to \citet{ruh-07}.
However, the voxel-based data term requires a huge, high-resolution intensity
volume, and a local window of $11^3$ voxels for matching, which lowers
spatial resolution, albeit less than earlier local matching.
\changes{
To overcome the memory bottleneck, we have further proposed a sparse particle representation \citep{las-18b}.
3D particle reconstruction and motion estimation are still performed sequentially.
Then local particle constellations are encoded in a descriptor and matched. The need to rely on spatial context limits the spatial resolution of the resulting flow field.
}
\citet{gre-14} proposed an approach similar to \citet{las-17} for dye-injected two-media
fluids. Their aim are visually pleasing, rather than physically
correct results, computed for relatively small volumes ($\approx
\!100^3$ voxels). We note that dye leads to data that is very different
from tracer particles, \changes{\ie,  %
it produces structures along the transition surface between water and dye that can be matched
well across time, but does not afford data evidence in large parts of the volume.}
\citet{dal-17} use compressive sensing to jointly recover
the location and motion of a sparse, time-varying signal with a mathematical recovery guarantee. Results are
only shown for small grids ($256^3$), and the physics of fluids is not
considered. %
\citet{bar-13} introduce a joint approach for
3D reconstruction and flow estimation, however, without considering
physical properties of the problem.
Their purely Eulerian, voxel-based setup limits the method to small
volume sizes, \ie, the method is only evaluated on a
$61\times61\times31$ grid and a rather low seeding density of
0.02 ppp.
\citet{xio-17} propose a joint formulation for
their single-camera PIV setup. The volume is illuminated by
rainbow-colored light planes that encode depth information. This
permits the use of only a single camera with the drawback of lower
depth accuracy and limited particle \changes{seeding} density.
Voxel occupancy probabilities are recovered on a 3D grid. To handle the
ill-conditioned problem from a single camera, constraints on particle
sparsity and motion consistency (including physical constraints) are
incorporated in the optimization. The method operates on a ``thin''
maximum volume of $512 \times 270 \times 20$. The single-camera setup
does not allow a direct comparison with standard 3D PIV/PTV, but can
certainly not compete in terms of depth resolution.
In contrast, by separating the representation of particles and motion
field, our hybrid Lagrangian/Eulerian approach allows for sub-pixel
accurate particle reconstruction and large fluid volumes.
Finally, \citet{ruh-05} propose a hybrid discrete particle and continuous variational motion estimation approach. 
Particle reconstruction and motion estimation are performed sequentially, 
and without physically plausible regularization of the flow. 

Volumetric fluid flow is also related to variational scene-flow
estimation, especially methods that parameterize the scene in 3D
space \citep{BashaMK2010Sceneflow3D,vogel11}.
Like those, we search for the geometry and motion of a dynamic scene
and exploit multiple views, yet our goal is a dense reconstruction in
a given volume, rather than a pixel-wise motion field.
Scene flow has undergone an evolution similar to the one for 3D fluid
flow. Early methods started with a fixed, precomputed geometry
estimate
\citep{Wedel2008SceneFlow2D,Rabe2010TemporalFilterSceneFlow2D}, with
a notable exception \citep{Huguet2007}. Later work moved to a joint
reconstruction of geometry and motion
\citep{BashaMK2010Sceneflow3D,ValgaertsBZWST10,vogel11}.
Likewise, \citet{els-06} and \citet{las-17} precompute the 3D tracer particles
before estimating their motion.
The method described in the present paper is, to our knowledge, the first multi-camera PIV scheme that
jointly determines the explicit locations of all particles and the physically constrained
motion of the fluid.

Several scene flow %
methods \citep{vogel13,Vogel2015,Menze2015CVPR} overcome the large
state space by sampling geometry and motion proposals, and perform discrete
optimization over those samples. In a similar spirit, we employ IPR to
generate discrete particle proposals, but then combine them with a
continuous, variational optimization scheme.
We note that discrete labeling does not suit our task: The
volumetric setting would require an excessive number of labels (3D
vectors), and enormous amounts of memory. And it does not lend itself
to sub-voxel accurate inference.

\changes{
The discretization of our motion field is based on the \textit{finite element method} (FEM) \citep{cou-43,red-93,bre-12}.
FEM has been applied to variational problems in computer vision, including 2D PIV \citep{ruh-06, ruh-07} and semantic 3D reconstruction \citep{ric-17}.
}

\section{Method}
\label{sec:method}

To set the scene, we restate the goal of our method: densely predict
the 3D flow field in a fluid seeded with tracer particles, from
multiple 2D views acquired at two adjacent time steps.

\begin{figure}[t]

 \centering
 \includegraphics[width=\columnwidth]{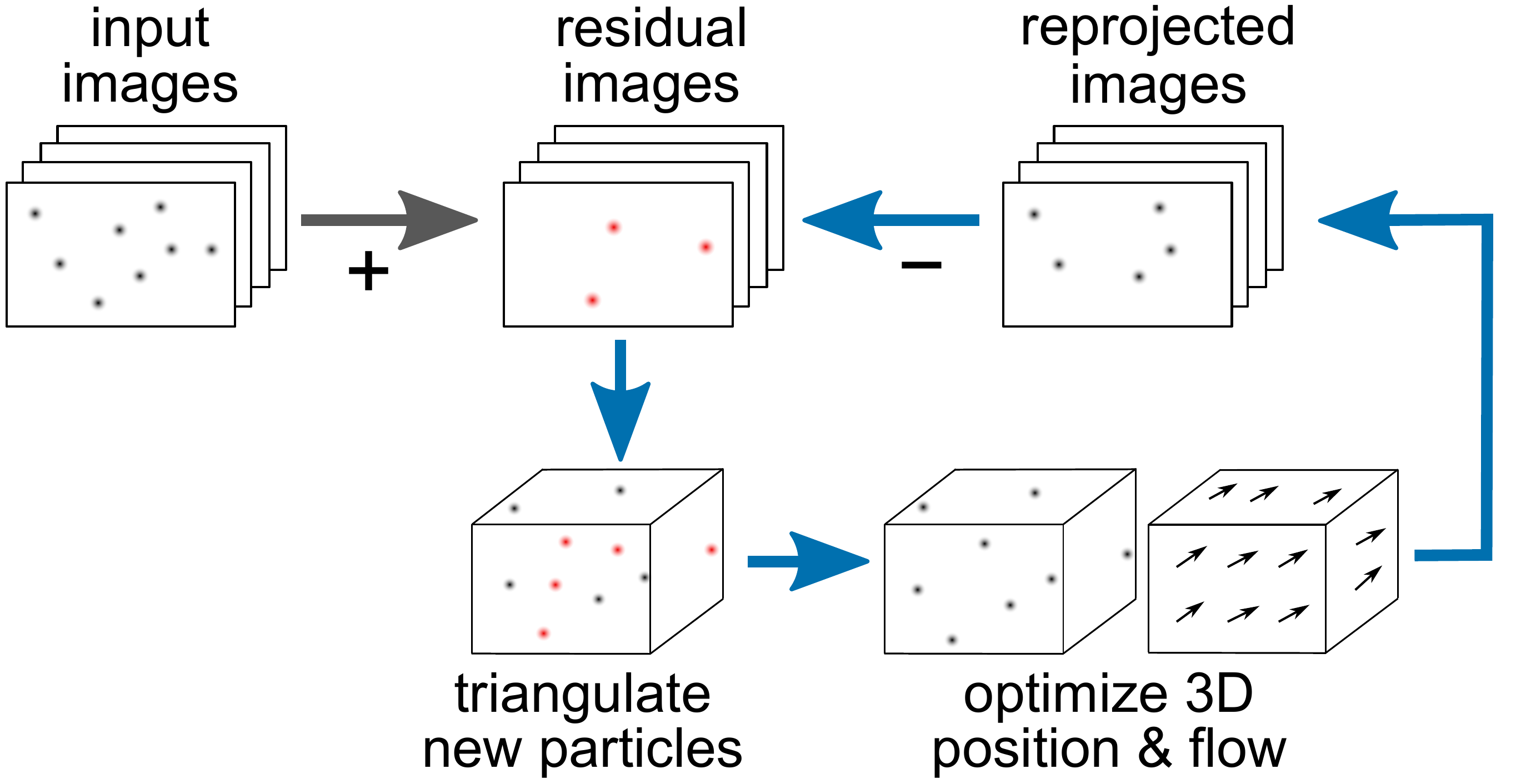}
 \caption{Particle position and flow estimation pipeline. We
 alternate between joint optimization of 3D particle positions and
 dense flow vectors, and adding
 new candidate particles by triangulation from the residual images.
   }
   \label{fig:pipeline}
    \vspace{-4mm}
\end{figure}

We aim to do this in a direct, integrated fashion. The joint particle
reconstruction and motion estimation is cast into a hybrid
Lagrangian/Eulerian model, where we recover individual particles
and keep track of their position and appearance, but reconstruct a
continuous 3D motion field in the entire domain.
A dense sampling of the motion field makes it technically
and numerically easier to adhere to physical constraints like
incompressibility.
In contrast, modeling particles explicitly takes advantage of the low
particle density in PIV data. Practical densities are around $0.1$
particles per pixel (ppp) in the images. Depending on the desired
voxel resolution, this corresponds to 10-1000 times lower volumetric
density.
Our complete pipeline is depicted in \Fig~\ref{fig:pipeline}.  It
alternates between generating particle proposals (\Sec\ref{sec:init})
based on the current residual images (starting from the raw input
images), and energy minimization to update all particles and motion
vectors (\Sec\ref{sec:optim}).
The corresponding variational energy function is described in \Sec\ref{sec:energy}.
In the process, particle locations and flow estimates are
progressively refined and provide a better initialization for further
particle proposals.

Particle triangulation is highly ambiguous, so the proposal generator
will inevitably introduce many spurious ``ghost'' particles
(\Fig~\ref{fig:partProp}).
A sparsity term in the energy reduces the influence of low intensity
particles that usually correspond to such ghosts, while true
particles, given the preliminary flow estimate, receive additional
support from the data of the second time step. In later iterations,
already reconstructed particles vanish in the residual images. This
allows for a refined localization of remaining particles,
as particle overlaps are resolved.

\paragraph{Notation and Preliminaries.}
The scene is observed by $K$ calibrated cameras $\mathcal{K}_k,
k=1,\ldots,K$, recording the images $\mathcal{I}^t_k$ at time $t$.
Parameterizing the scene with 3D entities obviates the need for image
rectification.
Fluid experiments typically need sophisticated models to deal with refraction (air-glass and glass-water), or an optical transfer function derived from volume self-calibration \citep{wie-08,sch-12}.
We keep the formulation general with a generic projection operator $\Pi_k$ per camera.
For convenience, we include two important cases of $\Pi_k$, the pinhole camera model (e.g.~for measurements in air) and a polynomial camera model designed for multi-media experiments (air-glass-water) \citep{sol-97}.

The dependency on time is denoted via superscript $t_0, t_1$, and
omitted when possible.
We denote the set of particles
$\mathcal{Q}:=\{ (p_i,c_i) \}_{i=1}^Q$,
composed of a set of intensities $\mathcal{C}:=\{ c_i \}_{i=1}^Q$,
$c_i\in\mathbb{R}^+_0$ and positions $\mathcal{P}:=\{ p_i \}_{i=1}^Q$, where
each $p_i\in\mathbb{R}^3$ is located in the rectangular domain
$\Omega\subset\mathbb{R}^3$.
The 3D motion field at position $x \in \Omega$, between times $t_0$
and $t_1$, is $\uf(x,\mathcal{U})$. The set $\mathcal{U}$
contains motion vectors $u\in\mathbb{R}^3$ located at a finite set of
positions $y\in\mathcal{Y}\subset\Omega$.
If we let these locations coincide with the particle positions, we
would arrive at a fully Lagrangian design, also referred to
as \emph{smoothed particle hydrodynamics}
\citep{Monaghan2005,Adams2007}.
In this work, we prefer a fixed set $\mathcal{Y}$ and represent
the functional $\uf(x,\mathcal{U})$ by trilinear interpolation, \ie we opt for
an Eulerian description of the motion. Our model is, thus, similar to
the so-called \emph{particle in cell design} \citep{Zhu2005}.
\changes{Without loss of generality,} we assume
$\mathcal{Y}\subset\Omega\cap\mathbb{Z}^3$, \ie, we set up a
regular grid of vertices $\mathbf{i}\in\mathcal{Y}$ of size
$N\times M \times L$, which induce a voxel covering
$V(\Omega)$ of size $N-1\times M-1 \times L-1$ of
the whole domain.
\changes{We use a trilinear FEM discretization, \ie,} each grid vertex
$\mathbf{i}=(\mathbf{i}_1,\mathbf{i}_2,\mathbf{i}_3)\trans$ is
associated with a trilinear basis function:

\begin{equation}\label{eq:basis}
 \begin{split}
\mathbf{b}_\mathbf{i}(x) :=
\prod_{l=1}^3\max(0,1-|&x_l-\mathbf{i}_l|), \\[-2ex]
&\textrm{ for } x=(x_1,x_2,x_3)\trans. %
  \end{split}
\end{equation}

The elements $u_{\mathbf{i}}\in\mathcal{U}$ now represent the
coefficients of our motion field function
$\uf(x,\mathcal{U}),x\in\Omega$
that is given by:
 \begin{equation}
 \begin{split}
   \label{eq:def_u}
  \uf(x,\mathcal{U}) =
  &\big(\uf_1(x,\mathcal{U}),\uf_2(x,\mathcal{U}),\uf_3(x,\mathcal{U})\big)\trans  \\[1ex] %
  &\textrm{with}\ \uf_l(x,\mathcal{U}) =
  \sum_{\mathbf{i}\in\mathcal{Y}} \mathbf{b}_\mathbf{i}(x) u_{\mathbf{i},l}\;,\ l=\textrm{1,2,3}.
  \end{split}
 \end{equation}
 
 \changes{
Finally, in our energy formulation we make use of the \emph{indicator function} 
$\delta_\mathcal{C}$ defined for a (not necessarily convex) set
$\mathcal{C}$ \citep{Boyd-2004, Bertsekas-99}:
\begin{equation}
\delta_\mathcal{C} (x) := 
\left\{
\begin{array}{cc} 0 &  \quad \textrm{ if } x \in \mathcal{C} \\
 						  \infty& \quad \textrm{else}
\end{array} 
\right..
\end{equation}
}
 
 \subsection{Energy Model}\label{sec:energy}
With the definitions above, we can write the energy
\begin{equation}\label{eq:Energy_functional}
E(\mathcal{P},\mathcal{C},\mathcal{U}) := \frac{1}{2} E_\textrm{D}(\mathcal{P},\mathcal{C},\mathcal{U}) +
\frac{\lambda}{2} E_\textrm{S}(\mathcal{U}) +
\mu E_\textrm{Sp}(\mathcal{C}),
\end{equation}
with a data term $E_\textrm{D}$, a smoothness term $E_\textrm{S}$
operating on the motion field, and a sparsity prior $E_\textrm{Sp}$
operating on the intensities of the particles.

\subsubsection{Data Term}
To compute the data term, the images of all cameras at both
time steps are predicted from the particles' positions and
intensities, and the 3D motion field.
$E_\textrm{D}$ penalizes deviations between predicted and
observed images:

\begin{equation}\label{eq:ED}
\begin{split}
&E_D(\mathcal{P}, \mathcal{C}, \mathcal{U}) :=
\\
&\;\frac{1}{K} \! \sum_{k=1}^K \!\int_{\Gamma_k}\!\!\! \big| \mathcal{I}^{t_0}_k(x) \! 
 -\!\!  \sum_{i=1}^P \!\Pi_k
( c_i\!\cdot\! \mathcal{N}(p_i,\sigma)(x) \! ) \big|_2^2 \dx \  +  \\
&\;\frac{1}{K} \! \sum_{k=1}^K \!\int_{\Gamma_k}\!\!\! \big| \mathcal{I}^{t_1}_k(x) \! 
 -\!\! \sum_{i=1}^P \!\Pi_k ( c_i\!\cdot\! \mathcal{N}(p_i\!+\!\uf(p_i,\mathcal{U}),\sigma)(x) \! ) \big|_2^2 \dx.
\end{split}
\end{equation}

Following an additive (in terms of particles) image formation model,
 we integrate over the image plane $\Gamma_k$ of camera $k$.
Individual particles $(p,c)\in\mathcal{Q}$ are represented by Gaussian
blobs with variance $\sigma^2$.
Particles do not exhibit strong shape or material variations. Their
distance to the light source does influence the observed intensity.
But since it changes smoothly and the cameras record with high
frame-rate, assuming constant intensity is a valid approximation for
our two-frame model.

In practical setups, the depth range of the volume $\Omega$ is small compared to the distance from the cameras. Therefore,
the projection is close to
orthographic, such that particles undergo almost no perspective distortion,
and
their image projections are 2D Gaussian blobs.
In that regime, and omitting constant terms, the expression for a
projected particle simplifies to
\begin{equation}\label{eq:ED_blob}
\begin{split}
\Pi \big( &\mathcal{N}(\cdot,\sigma)(x) \big) \approx \\
&\mathcal{N}\big(\Pi(\cdot),\sigma\big)(x)
\propto \sigma^{-1}\exp
\big(-| \Pi(\cdot)-x |^2\sigma^{-2}\big).
\end{split}
\end{equation}

When computing the derivatives of \eqref{eq:ED} \wrt the
set of parameters, we do not integrate the particle blobs over
the whole image, but restrict the area of influence of \eqref{eq:ED_blob}
to a radius of $3\sigma$, covering $99.7$\% of its total intensity.

\paragraph{Camera Model.}

For measurements in air a simple pinhole camera model is sufficient to model the camera geometry. However, when performing experiments in water, cameras are positioned outside of the water tank. Hence, light gets refracted at air-glass and glass-water transitions. To model this complex setup, \citet{sol-97} proposed a polynomial camera model which is commonly used for 3D-PIV/PTV measurements.
Both proposed camera models are special cases of the cubic rational polynomial camera model \citep{har-97}.

The projection operator $\Pi_K(p_i)$ maps 3D particle locations $p_i$ to 2D camera coordinates $(u_{K,i},v_{K,i})$. We omit the subscript $K$ per camera and the particle index $i$ in the following for better readability.
For the pinhole camera model the mapping is defined as follows:
\begin{equation}
u = \frac{P^{1T} p}{P^{3T} p},\quad v = \frac{P^{2T} p}{P^{3T} p}.
\end{equation}
For the polynomial camera model by \citet{sol-97}, 38 parameters $a_i \in \mathbb{R}^2, i=0\ldots,18 $ are needed to model the cubic dependencies in $x$ and $y$ direction and the quadratic dependency in $z$ direction (assuming the thinnest extend in $z$ direction). Also note that the perspective division is omitted, which is possible due to the specific 3D-PIV/PTV setup with thin measurement volumes:

\begin{equation}
\begin{split}
(u,v) &= a_0  + a_1 p_x + a_2 p_y+ a_3 p_z 
+ a_4 p_x^2 + a_5 p_x p_y \\ 
&+ a_6 p_y^2
+a_7 p_x p_z + a_8 p_y p_z + a_9 p_z^2  
+ a_{10} p_x^3  \\ 
&+  a_{11} p_x^2 p_y + a_{12} p_x p_y^2
+ a_{13} p_y^3 + a_{14} p_x^2 p_z \\
& + a_{15} p_x p_y p_z 
+ a_{16} p_y^2 p_z +   a_{17} p_x p_z^2 + a_{18} p_y p_z^2.
\end{split}
\end{equation}

\subsubsection{Sparsity Term}
The majority of the generated candidate particles do not correspond to
\emph{true} particles.
To suppress the influence of the large set of low-intensity ghost
particles one can exploit the expected sparsity of the
solution, \eg \cite{Petra2009}. In other words, we aim to reconstruct
the observed scenes with \emph{few, bright} \changes{particles},
by introducing the following energy term:
\vspace{-0.25em}
\begin{equation}\label{eq:ESP}
E_\textrm{Sp}(\mathcal{C}):= \sum_{i=1}^Q |c_i|_\diamond + \delta_{\{\geq0\}} (c_i).
\end{equation}
Here, $\delta_{\Delta}(\cdot)$ denotes the indicator function of the
set $\Delta$. Note, this term additionally excludes negative intensities.
Although not directly related to sparsity, we identified \eqref{eq:ESP}
as a convenient place to include this constraint.
Popular sparsity-inducing norms $|\cdot|_\diamond$ are either the $1$- or
$0$-norm ($|\cdot|_\diamond=|\cdot|_1$, respectively
$|\cdot|_\diamond=|\cdot|_0$).
We have investigated both choices and prefer the stricter $0$-norm for
the final model.
The $0$-norm counts the number of non-zero intensities and rapidly
discards particles that fall below a certain intensity threshold
(modulated by $\mu$ in \eqref{eq:Energy_functional}). While the
$1$-norm only gradually reduces the intensities of weakly
supported particles.

\subsubsection{Smoothness Term}
As in our previous work \citep{las-17} we
employ a quadratic regularizer per component of the flow gradient
and a term that enforces a divergence-free motion field
to define a suitable smoothness prior:
\begin{equation}\label{eq:ES}
E_\textrm{S}(\mathcal{U}):=\int_\Omega \sum_{l=1}^3|\nabla \uf_l(x,\mathcal{U})|_2^2 + \delta_{\{0\}}\big(\div\uf(x,\mathcal{U})\big) \dx.
\end{equation}
In \citep{las-17} we have shown that \eqref{eq:ES} has a physical
interpretation, in that the stationary Stokes equations emerge as the
Euler-Lagrange equations of the energy \eqref{eq:ES}, including an
additional force field.
Thus, \eqref{eq:ES} models the incompressibility of the fluid, while
$\lambda$ represents its viscosity.
In \citep{las-17} we also suggest a variant in which the
 hard divergence constraint is replaced with a soft penalty:
\begin{equation}\label{eq:ES_soft}
E_{\textrm{S},\alpha}(\mathcal{U}):=\int_\Omega \sum_{l=1}^3|\nabla \uf_l(x,\mathcal{U})|_2^2 + \alpha | \div\uf(x,\mathcal{U})|^2 \dx.
\end{equation}
This version simplifies the numerical optimization, trading off speed
for accuracy. For adequate (large) $\alpha$, the results are similar
to the hard constraint in \eqref{eq:ES}.
\Eq \eqref{eq:ES} requires the computation of divergence $\div$ and gradients
$\nabla$ of the 3D motion field.
Following the definition \eqref{eq:def_u} of the flow field, both entities
are linear in the coefficients $\mathcal{U}$ and constant per
voxel $v\in V(\Omega)$.
A valid discretization of the divergence operator can be achieved via
the divergence theorem:
\begin{equation}\label{eq:discretisation}
\begin{split}
&
\int_v \div u(x)\dx = 
\int_{\partial v}  \langle \mathbf{\nu}(x), u(x)\rangle \dx = \\
& \sum_{\mathbf{i}}  %
\int_{\partial v} \!\!
b_\mathbf{i}(x)
\langle \mathbf{\nu}(x), u_{\mathbf{i}} \rangle \dx =
\frac{1}{4}
\sum_{l=1}^3
\sum_{(\mathbf{i}, \mathbf{j}) \in \mathcal{Y}\cap v:\mathbf{i}-\mathbf{j}=e_l}
\!\!\!\!\!\!\!\!\!\!\!
u_{\mathbf{i},l}-u_{\mathbf{j},l},
\end{split}
\end{equation}
where we let $\nu(x)$ denote the outward-pointing normal of voxel $v$ at
position $x \in v$ and $e_l$ the unit vector in direction $l$.
The final sum considers pairs of corner vertices
$(\mathbf{i},\mathbf{j})\in\mathcal{Y}\cap v$ of voxel $v$,
adjacent in direction $l$.
The definition of the per-voxel gradient follows from \eqref{eq:def_u}
in a similar manner.

\changes{
Our approach ressembles that of conforming FEM, with a trilinear basis for the
velocity field and a per-voxel constant basis for the dual functions (constant pressure per voxel). 
Consequently, for velocity fields contained in the trilinear subspace of functions 
\Eq~\eqref{eq:discretisation} computes the divergence for every point 
of the continuous domain contained in the interior of voxel $v$. 
Despite being popular and in many applications adequate \citep{lan-2002}, 
our pair of trial and test FEM spaces does not fulfill the 
Babuska-Brezzi condition \citep{bre-12}.
In our experiments, we did not observe any artifacts in the estimated flow fields, however, 
within our framework it is straightforward to apply 
various known stabilization techniques \citep{lan-2002}, 
or to switch to a FEM representation that does satify the condition,
for instance 
the Taylor-Hood element \citep{tay-73}.
}

\subsection{Particle Initialization}\label{sec:init}
To find putative particles, we employ a direct detect-and-triangulate strategy
like IPR \citep{wie-13}.
Having found an initial set of particles, we minimize the energy \eqref{eq:Energy_functional}, reproject the reconstructed 3D particles, compute residual images, and rerun the particle detection to find additional particles. This alternation scheme continues until no new particles are found \changes{(\cf \Fig~\ref{fig:pipeline})}.

\begin{figure}[t]
 \centering
\includegraphics[width=\columnwidth]{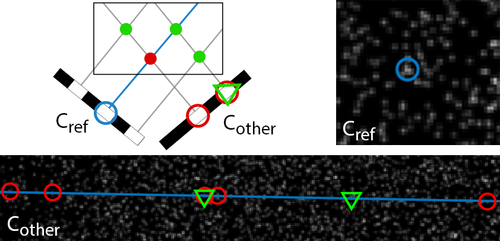}
    \vspace{-4mm}
 \caption{
  A particle in the reference camera (\textcolor{blue}{circle}) can lead to multiple epipolar-consistent putative matches (\textcolor{red}{circles}). However, only a subset of them represents true 3D particles (\textcolor{green}{triangles}).
 \emph{Left}: 1D-illustration. \emph{Right}: peak in reference camera (0.1ppp). \emph{Bottom:} other camera view with 5 putative matches that are consistent over all 4 cameras. 
 Moreover, one true particle has initially not been found, due to
visually overlapping ones that bias peak detection; but will be
found in subsequent iterations, when the overlapping particles have been
reconstructed and removed from the residual images or the triangulation threshold has been relaxed.
}

   \label{fig:partProp}

\end{figure}

Particle triangulation is extremely ambiguous and not decidable with
local cues (\Fig~\ref{fig:partProp}). Instead, \emph{all} plausible
correspondences are instantiated.
One can interpret the process as a proposal generator for the set of particles,
which interacts with the sparsity constraint \eqref{eq:ESP}.
This proposal generator creates new candidate particles where image
evidence remains unexplained.
The sparsity prior ensures that only ``good'' particles survive and contribute
to the data costs; whereas those of low intensity, which do not contribute to lowering the energy, fade to
``zero-intensity'' particles
(particles of very low intensity are uncommon in reality).
After each iteration the zero-intensity particles are actively
discarded from $\mathcal{Q}$ to reduce the workload.
Note that this does not change the energy of the current solution.
After the first particles and a coarse motion field have been
reconstructed, a better initialization is available to spawn further
particles, in locations suggested by the residual maps between
predicted and observed images.
Particles that contribute to the data are retained in the subsequent
optimization and help to refine the motion field, \etc

The procedure is inspired by the heuristic, yet highly efficient,
iterative approach of \citep{wie-13}. They also refine particle
candidates triangulated from residual images.
Other than theirs, our updated particle locations follow from a joint
spatio-temporal objective, and thus also integrate information from
the second time step.

In more detail, each round of triangulation proceeds as follows:
first, detect peaks in 2D image space for all cameras at time step
$t_0$. In the first iteration this is done in the raw inputs, then in
the residual images
$\mathcal{I}_{k,\textrm{res}}^{t_0} := \int_{\Gamma_k} \mathcal{I}^{t_0}_k(x) - \sum_{i=1}^Q \Pi_k
(c_i \cdot \mathcal{N}\big(p_i,\sigma)(x)\big) \dx$.
Peaks are found by non-maximum suppression with a $3\times3$
kernel, followed by sub-pixel refinement of all peaks with
intensity above a threshold $I_{min}$.
We treat one of the cameras, $k=1$, as reference and compute the
entry and exit points to $\Omega$ for a ray passing through each peak.
Reprojecting the entry and exit into other views yields epipolar line
segments, along which we scan for (putatively) matching peaks
(\Fig\ref{fig:partProp}).
Whenever we find peaks in all views that can be triangulated with a
reprojection error below a tolerance $\epsilon$, we generate a new
candidate particle.
Its initial intensity is set as a function of the intensity in the reference
view and the number of candidates:
if $m$ proposals $p_i$ are generated at a peak in the reference image,
we set $c_i:=\mathcal{I}_1\big(\Pi_1(p_i)\big) K / (K-1+m)$ for each of them.

\subsection{Energy Minimization}\label{sec:optim}
Our optimization is embedded in a two-fold coarse-to-fine scheme.
On the one hand, we start with a larger value for $\sigma$, so as to
increase the particles' basins of attraction and improve convergence.
During optimization, we progressively reduce $\sigma$ until we reach
$\sigma=1$, meaning that a particle blob covers approximately the
same area as in the input images.
On the other hand, we also start at a coarser grid $\mathcal{Y}$ and
refine the grid resolution along with $\sigma$.

To minimize the non-convex and non-smooth
energy \eqref{eq:Energy_functional} for a given $\sigma$, we employ
PALM \citep{Bolte2014}, %
in its inertial variant \citep{poc-16}.
Because our energy function is
semi-algebraic \citep{Bolte2014}, it satisfies the
Kurdyka-Lojasiewicz property \citep{Bolte2006}, therefore the sequence
generated by PALM globally converges to a critical point of the energy.
The key idea of PALM is to split the variables into blocks, such that
the problem is decomposed into one smooth function on the entire
variable set, and a sum of non-smooth functions in which each block is
treated separately.
We start by arranging the locations and intensities of the particles
$\mathcal{Q}$ into two separate vectors
$\mathbf{p}:=(p_1\trans,\ldots,p_Q\trans)\trans\in\mathbb{R}^{3Q}$
and
$\mathbf{c}:=(c_1,\ldots,c_Q)\trans\in\mathbb{R}^{Q}$. Similarly,
we stack the coefficients of the trilinear basis
$\mathbf{u}:=(u_{\mathbf{i},1}\trans, u_{\mathbf{i},2}\trans,
u_{\mathbf{i},3}\trans)_{\mathbf{i}\in\mathcal{Y}}\trans\in\mathbb{R}^{3NML}$.
With these groups, we split the energy functional into a smooth
part $H$ and two non-smooth functions, $F_\mathbf{c}$ for the intensities $\mathbf{c}$
and $F_\mathbf{u}$ for the motion vectors $\mathbf{u}$:
\begin{equation}\label{eq:energy_opt}
\begin{split}
&
E(\mathbf{p}, \mathbf{c}, \mathbf{u}):= H(\mathbf{p}, \mathbf{c}, \mathbf{u})
+ F_\mathbf{c}(\mathbf{c}) + F_\mathbf{u}(\mathbf{u}) + F_\mathbf{p}(\mathbf{p}), \\
&\textrm{with} \\
&H(\mathbf{p}, \mathbf{c}, \mathbf{u}):= E_D(\mathbf{p}, \mathbf{c}, \mathbf{u}) + \lambda
\sum\nolimits_{l=1}^3 \| \nabla\mathbf{u}_l \|^2,\\
&F_\mathbf{c}(\mathbf{c}) := \mu E_{Sp}(\mathbf{c}), \\ %
&F_\mathbf{u}(\mathbf{u}) := \delta_{\{0\}}(\div\mathbf{u}),\\
&F_\mathbf{p}(\mathbf{p}) := 0.
\end{split}
\end{equation}
For notation convenience, we define $F_\mathbf{p}(\mathbf{p}):=0$.
The algorithm then alternates the steps of a proximal forward-backward
scheme: take an explicit step \wrt one block of variables
$z\in\{\mathbf{p},\mathbf{c},\mathbf{u}\}$ on the smooth part $H$
of the energy function, then take a backward (proximal) step on the
non-smooth part $F_z$ \wrt the same variables.
That is, we alternate steps of the form
\begin{equation}\label{eq:PFB_step}
\begin{split}
z^{n+1} =\ &\textrm{prox}^{F_z}_t(z) := \argmin_y F_z(y) + \frac{t}{2} \| y-z \|^2, \\
&\textrm{ with } z = z^n - \frac{1}{t} \nabla_z H(\cdot,z^n,\cdot),
\end{split}
\end{equation}
with a suitable step size $1/t$ for each block of variables.
Here and in the following, the placeholder variable $z$ can stand for
$\mathbf{c}$, $\mathbf{p}$ or $\mathbf{u}$, as required.

A key property is that, throughout the iterations, the partial
gradient of function $H$ \wrt a variable block
$z\in\{\mathbf{p},\mathbf{c},\mathbf{u}\}$ must be globally
Lipschitz-continuous with some modulus $L_z$ at the current solution:
\begin{equation}\label{eq:Lipshitz_condition}
\|\nabla_z H(\cdot,z_1,\cdot) - \nabla_z H(\cdot,z_2,\cdot)\| \leq L_z(\cdot,\cdot) \|z_1-z_2\| \,\forall z_1,z_2.
\end{equation}
In other words, before we accept an update $z^{n+1}$ computed
with \eqref{eq:PFB_step}, we need to verify that the step size $t$
in \eqref{eq:PFB_step} fulfills the descent
lemma \citep{Bertsekas1989}: %
\begin{equation}\label{eq:Lipshitz_test}
\begin{split}
E( \cdot, z^{n+1}, \cdot ) \leq & E( \cdot, z^n, \cdot ) +
\langle \nabla_z H(\cdot, z^n, \cdot), z^{n+1}-z^{n}\rangle \\
& + \frac{t}{2} \|z^{n+1}-z^n\|^2.
\end{split}
\end{equation}
Note that Lipschitz continuity of the gradient of $H$
need not be tested globally, but can be verified locally at the current solution.
This property allows for a back-tracking approach to determine the
Lipschitz constant, \eg \citet{Beck2009}. %
Algorithm~\ref{alg:ipalm} provides pseudo-code for our scheme to minimize
the energy \eqref{eq:energy_opt}.
To accelerate convergence we apply extrapolation (lines 4/8/12).
These inertial steps, \cf \citet{poc-16,Beck2009}, significantly reduce
the number of iterations in the algorithm,
while leaving the computational cost per step practically untouched.
We also found it useful to not only reduce the step sizes (lines 7/11/15 %
in \Alg\ref{alg:ipalm}), but also to increase them, as long
as \eqref{eq:Lipshitz_test} is fulfilled,
so as to make the steps per iteration as large as possible.

\begin{algorithm}[tb]
\caption{iPalm implementation for energy \eqref{eq:energy_opt}}\label{alg:ipalm}
\begin{algorithmic}[1]
\Procedure{ipalm}{$\mathbf{p}^0,\mathbf{c}^0, \mathbf{u}^0$}
\State
$
\mathbf{p}^{-1} \gets \mathbf{p}^{0};
\mathbf{c}^{-1} \gets \mathbf{c}^{0};
\mathbf{u}^{-1} \gets \mathbf{u}^{0};$
\State
$\tau\gets\frac{1}{\sqrt{2}};
L_\mathbf{p}\gets 1;L_\mathbf{c}\gets 1;L_\mathbf{u}\gets 1;$
\For{n:=0 to $n_\textrm{steps}$ and while not converged}
\State $\mathbf{\hat{p}} \gets \mathbf{p}^n + \tau (\mathbf{p}^n-\mathbf{p}^{n-1}) ;$ // inertial step
\While{true}
\State
$
\mathbf{p}^{n+1} := \mathbf{\hat{p}} - 1/L_\mathbf{p} \nabla_\mathbf{p} H(\mathbf{\hat{p}}, \mathbf{c}^n, \mathbf{u}^n)
$;
\If{$\mathbf{p}^{n+1}, L_\mathbf{p}$ fulfill \eqref{eq:Lipshitz_test}}
{\textbf{break};}
\Else{ $L_\mathbf{p}=2 L_\mathbf{p}$};%
\EndIf
\EndWhile
\State $\mathbf{\hat{c}} \gets \mathbf{c}^n + \tau (\mathbf{c}^n-\mathbf{c}^{n-1}) $; // inertial step
\While{true}
\State
$
\mathbf{c} := \mathbf{\hat{c}} - 1/L_\mathbf{c} \nabla_\mathbf{c} H(\mathbf{p}^{n+1}, \mathbf{\hat{c}}, \mathbf{u}^n);
$
\State
$
\mathbf{c}^{n+1}:= \textrm{prox}^{F_\mathbf{c}}_{L_\mathbf{c}}(\mathbf{c})$; // \Eq\eqref{eq:PFB_step}
\If{$\mathbf{c}^{n+1}, L_\mathbf{c}$ fulfill \eqref{eq:Lipshitz_test}}
{\textbf{break};}
\Else{ $L_\mathbf{c}=2 L_\mathbf{c}$;} %
\EndIf
\EndWhile
\State $\mathbf{\hat{u}} \gets \mathbf{u}^n + \tau (\mathbf{u}^n-\mathbf{u}^{n-1}) $; // inertial step
\While{true}
\State
$
\mathbf{u} := \mathbf{\hat{u}} - 1/L_\mathbf{u} \nabla_\mathbf{u} H(\mathbf{p}^{n+1}, \mathbf{c}^{n+1}, \mathbf{\hat{u}})
$;
\State
$
\mathbf{u}^{n+1}:= \textrm{prox}^{F_\mathbf{u}}_{L_\mathbf{u}}(\mathbf{u})$; // \Eq\eqref{eq:PFB_step}
\If{$\mathbf{u}^{n+1}, L_\mathbf{u}$ fulfill \eqref{eq:Lipshitz_test}}
{\textbf{break};}
\Else{ $L_\mathbf{u}=2 L_\mathbf{u}$;}
\EndIf
\EndWhile
\EndFor
\EndProcedure
\end{algorithmic}
\afterpage{\global\setlength{\textfloatsep}{8pt}}%
\end{algorithm}

One last thing needs to be explained, namely how we find the solution
of the proximal steps on the intensities $\mathbf{c}$ and flow vectors
$\mathbf{u}$.
The former can be solved point-wise, leading to the following
1D-problem:
\begin{equation}
\textrm{prox}^{F_c}_t (\bar{c}) := \argmin_c \mu |c|_\diamond + \delta_{\{\geq0\}} (c) + \frac{t}{2} | c-\bar{c} |^2,
\end{equation}
which admits for a closed-form solution for both %
the $0$-norm $|c|_0$ and the $1$-norm $|c|_1$:
\begin{equation}\label{eq:prox_c}
\textrm{prox}^{|\cdot|_0}_t (\bar{c}) :=
\left\{
   \begin{array}{c@{\hspace{0.2cm}}l}
     0          & \textrm{if } t \bar{c}^2 < 2\mu \textrm{ or } \bar{c}<0\\
     \bar{c}    & \textrm{else},
   \end{array}
   \right.,
\end{equation}
\begin{equation}
\textrm{prox}^{|\cdot|_1}_t (\bar{c}) := \max(0,\bar{c} - \mu/t).
\end{equation}

The proximal step for the flow vector,
$
\textrm{prox}^{F_\mathbf{u}}_t (\mathbf{\bar{u}}) := \argmin_\mathbf{u} \delta_{\{0\}}(\div\mathbf{u}) + \frac{t}{2} \| \mathbf{u}-\mathbf{\bar{u}} \|^2
$,
requires the projection of $\mathbf{\bar{u}}$ onto the space of
divergence-free 3D vector fields.
Hence,
given $\mathbf{\bar{u}}$, the solution is independent of the step size $1/t$,
which we omit in the following.
We construct the Lagrangian by introducing the multiplier
$\mathbf{\phi}$, a scalar vector field whose physical meaning is the
pressure in the fluid \citep{las-17}:
\begin{equation}
\min_\mathbf{u}\max_\mathbf{\phi}\frac{1}{2}\|\mathbf{u}-\mathbf{\bar{u}}\|^2 + \mathbf{\phi}\trans \nabla\cdot \mathbf{u}.
\end{equation}
To prevent confusion, we introduce $D\mathbf{u}$ as matrix notation
for the linear divergence operator $(\div\mathbf{u})$
in \eqref{eq:discretisation}.
The KKT conditions of the Lagrangian yield a linear equation
system. Simplification with the Schur complement leads to a Poisson
system, which we solve for the pressure $\mathbf{\phi}$ to get the
divergence-free solution:
\begin{equation}\label{eq:prox_u}
\textrm{prox}^{F_\mathbf{u}}_t (\mathbf{\bar{u}}) := \mathbf{\bar{u}} - D\trans \mathbf{\phi}
\quad\textrm{with }\;
D\trans D \mathbf{\phi} = D\trans \mathbf{\bar{u}}.
\end{equation}
Again interpreted physically, the divergence of the motion field is
removed by subtracting the gradient of the resulting pressure field.
For our problem of fluid flow estimation, it is not necessary to
exactly solve the Poisson system in every iteration. Instead, we keep
track of the pressure field $\mathbf{\phi}$ during optimization, and
warm-start the proximal step. In this way, a few (10-20) iterations of
preconditioned conjugate gradient descent suffice to update
$\mathbf{\phi}$.

If we replace the hard divergence constraint with the soft penalty
$E_{S,\alpha}$ from \eqref{eq:ES_soft}, we add $E_{S,\alpha}$ to the
smooth function $H$ in \eqref{eq:energy_opt}. Then only the proximal
step on the intensities $\mathbf{c}$ is needed in \Alg\ref{alg:ipalm}.
We conclude by noting that
accelerating the projection step is in itself an active
research area in fluid simulation
\citep{Ladicky2015,TompsonSSP16}.

\section{Evaluation}
\label{sec:evaluation}

Since there is no other measurement technique that could deliver ground
truth for fluid flow, 
we follow the standard practice and generate datasets for quantitative
evaluation via \emph{direct numerical simulations} (DNS) of turbulent
flow, using the \emph{Johns Hopkins Turbulence Database}
(JHTDB)~\citep{li-08,per-07}. This allows us to render realistic input
images with varying particle densities and flow magnitudes, together
with ground truth vectors on a regular grid.
We evaluate how our approach performs with different smoothness terms, particle
densities, initialization methods, particle sizes and temporal sampling rates.
Additionally, we show results on ``test case D'' of the
4$^\text{th}$ International PIV Challenge \citep{kah-16} and
quantitatively compare to the best performing method \citep{sch-16}.

 \begin{figure}[tb]
 \centering
 \includegraphics[width=0.32\columnwidth]{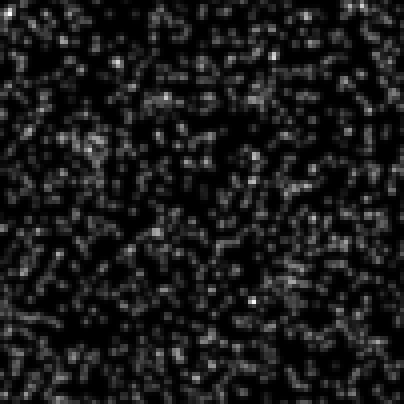}
 \includegraphics[width=0.32\columnwidth]{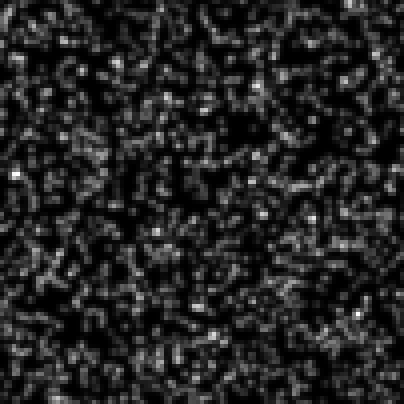}
 \includegraphics[width=0.32\columnwidth]{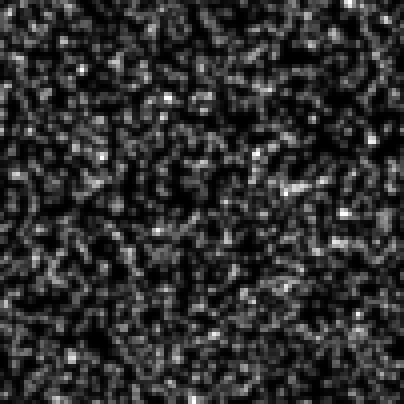}
    \vspace{-2mm}
 \caption{
Detail of rendered 2D particle images for particle densities of 0.125, 0.175
and 0.225 ppp (from left to right).
 }
 \label{fig:ppp}
\end{figure}

\paragraph{Simulated dataset.}
We follow the guidelines of the 4$^\text{th}$ International PIV
Challenge \citep{kah-16} for the setup of our own dataset: Randomly
sampled particles are rendered to four symmetric cameras of resolution
$1500\times800$ pixels, with viewing angles of $\pm
35^{\circ}$ \wrt the $yz$-plane of the volume, respectively $\pm
18^{\circ}$ \wrt the $xz$-plane. If not specified otherwise, particles
are rendered as Gaussian blobs with $\sigma=1$ and varying
intensity.
We sample 12 datasets from 6 non-overlapping spatial and 2 temporal 
locations ($t = 0.364$ and $t=8.364$) of the forced isotropic turbulence simulation of
the JHTDB. 
Discretizing each DNS grid point with 4 voxels, identical to \citet{kah-16}, 
each dataset corresponds to a volume size of $1024\times512\times352$.
Standard temporal difference between two consecutive time steps is $\Delta t = 0.004$.
For each dataset we sample $480,000$ seeding particles at random locations within the volume and ground truth flow vectors at each DNS grid location.
A subset of those particles is rendered on the actual camera views depending on the desired particle seeding density. In \Fig\ref{fig:ppp} we show $100 \times 100$ patches of rendered camera views with seeding densities of $0.125$, $0.175$ and $0.225$ ppp. Note that with higher seeding densities particle occlusions and overlaps increase.
For our flow fields with flow magnitudes up to $8.8$ voxels, we use
$10$ pyramid levels with downsampling factor $0.94$.
At every level we alternate between triangulation of candidate
particles and minimization of the energy function (at most $40$
iteration per level).

The effective resolution of the reconstructed flow field is determined
by the particle density. At a standard density of $\approx 0.1$ ppp
and a depth range of $352$ voxels, we get a density of $\approx
0.0003$ particles per voxel. This suggests to estimate the flow on a
coarser grid.
We empirically found a particle density of $0.3$ per voxel to still deliver
good results. Hence, we operate on a subsampled voxel grid of 10-times lower resolution per dimension in all our experiments, to achieve a notable speed-up and memory saving.
The computed flow is then upsampled to the target resolution, with barely any loss of accuracy.

We always require a 2D intensity peak in all four cameras to
instantiate a candidate particle. We start with %
a strict threshold of $\epsilon=0.8$ for the triangulation error,
as suggested in \citep{wie-13}, which is relaxed to
$\epsilon=2.0$ in later iterations.
The idea is to first recover particles with strong support, and
gradually add more ambiguous ones, as the residual images become
less cluttered.
We set $\lambda=0.04$ for our dataset.
Since $\lambda$
corresponds to the viscosity it should be adapted for other fluids.
We empirically set the sparsity weight $\mu=0.0001$.

\begin{table}[t]
\begin{center}
  \caption{
  Endpoint error (AEE), angular error (AAE) and absolute
  divergence (AAD) for different regularizers  (0.1 ppp).
}\label{tab:regulariser}
\vspace{\tablecaptionspace}
\renewcommand{\arraystretch}{1}
\newcolumntype{L}[1]{>{\raggedright\let\newline\\\arraybackslash\hspace{0pt}}m{#1}}
\begin{tabular}{L{1.0cm}cccc}
\hline
\rowcolor{gray!10}
 & $E_\textrm{S}$ &  $E_{\textrm{S},64}$ &  $E_{\textrm{S},0}$ & \citet{las-17} \\
\hline
AEE & 0.136 & 0.135 & 0.157  & 0.406  \\
AAE & 2.486 & 2.463 & 2.870 & 6.742 \\
AAD & 0.001 & 0.008 & 0.100 & 0.001 \\
\hline
\end{tabular}
\end{center}
\end{table}

\begin{figure}
\centering

 \includegraphics[width=\columnwidth]{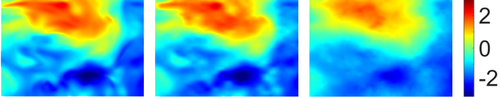}
    \vspace{-4mm}
 \caption{ Detail from an $xy$-slice of the flow in
 $X$-direction. \emph{Left to right:} Ground truth, our method and
 result of \citet{las-17}.
 } \label{fig:results}
\end{figure}

\begin{table*}[tb]
\begin{center}
  \caption{ Influence of particle density on our joint approach, as
 well as several baselines.}
\label{tab:ppp}
\vspace{\tablecaptionspace}

\begin{tabular}{c|ccc|ccc|ccc|ccc}
\hline
\rowcolor{gray!10}
ppp & \multicolumn{3}{c|}{\textbf{IPR joint}} & \multicolumn{3}{c|}{IPR sequential} & \multicolumn{3}{c|}{MART} & \multicolumn{3}{c}{true particles}  \\
\rowcolor{gray!10}
& AEE & \textit{prec.} & \textit{recall} & AEE & \textit{prec.} & \textit{recall} & AEE & \textit{prec.} & \textit{recall} & AEE & \textit{prec.} & \textit{recall} \\
\hline
0.1 & 0.136 & 99.98 & 99.95  & 0.136 & 99.97 & 99.96 & 0.232 & 70.39 & 83.93 & 0.136 & 100 & 100 \\
0.125 & 0.124 & 91.00 & 99.95 & 0.157 & 61.55 & 97.55 & 0.270 & 48.51 & 73.83 & 0.125 & 100 & 100  \\
0.15 & 0.115 & 82.82 & 99.95 & 0.310 & 33.46 & 85.09 & 0.323 & 44.61 & 70.17 & 0.118 & 100 & 100 \\
0.175 & 0.111 & 71.37 & 99.93 & 0.332 & 26.63 & 71.07 & 0.385 & 40.89 & 65.29 & 0.110 & 100 & 100 \\
0.2 & 0.108 & 55.43 & 99.86 & 0.407 & 19.42 & 64.26 & 0.506 & 36.88 & 58.13 & 0.105 & 100 & 100 \\
0.225  & 0.113 & 34.39 & 99.87 & & & & & & & 0.101 & 100 & 100 \\
0.25 & 0.134 & 24.54 & 99.51 & & & & & & & 0.098 & 100 & 100 \\

\hline
\end{tabular}
\end{center}
\end{table*}

\paragraph{Regularization.}
Our framework allows us to plug in different smoothness terms.
We show results for hard ($E_\textrm{S}$)
and soft divergence regularization ($E_{\textrm{S},\alpha}$).
\emph{Average endpoint error} (AEE), \emph{average angular error} (AAE), and \emph{average
absolute divergence} (AAD) are displayed in \Tab~\ref{tab:regulariser}.
Compared to our default regularizer $E_\textrm{S}$,
removing the divergence constraint ($\alpha=0$),
increases the error by $\approx15\%$.
With the soft constraint at high $\alpha=64$, the results
are equal to those of $E_\textrm{S}$.
We also compare to our previous sequential Eulerian-based approach \citep{las-17}.
Our joint model
improves the performance by $\approx70\%$ over that recent baseline,
on both error metrics.
In \Fig \ref{fig:results} we visually compare our results (with
hard divergence constraint) to those of \citet{las-17}.
The figure shows the flow in $X$-direction in one particular
$xy$-slice of the volume. Our method recovers %
a lot finer details, and is clearly closer to the ground truth.

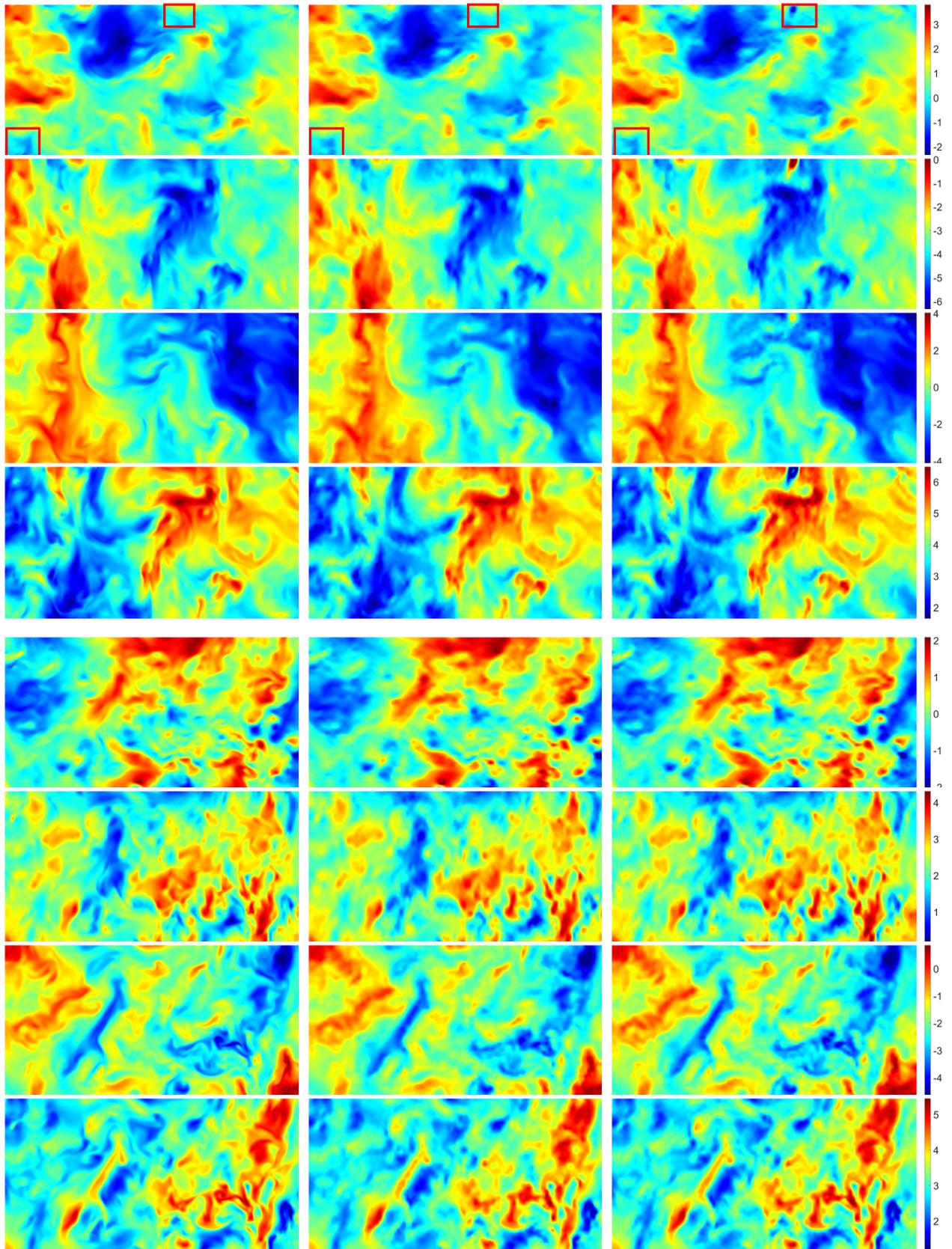
\begin{figure*}
\vspace{-3mm}
\centering
\begin{tikzpicture}
    \node[anchor=south west,inner sep=0] (image) at (0,0) {\includegraphics[trim={0 0 2cm 0},clip,height=3cm]{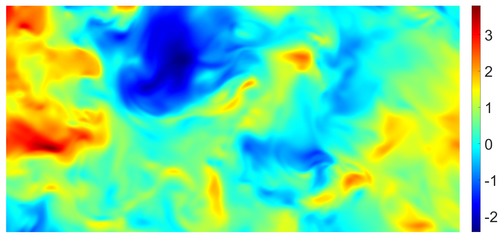}};
    \begin{scope}[x={(image.south east)},y={(image.north west)}]
        \draw[red, very thick] (0.55,0.85) rectangle (0.65,0.98);
        \draw[red, very thick] (0.02,0.08) rectangle (0.13,0.25);
    \end{scope}
    \end{tikzpicture}
    \begin{tikzpicture}
    \node[anchor=south west,inner sep=0] (image) at (0,0) {\includegraphics[trim={0 0 2cm 0},clip,height=3cm]{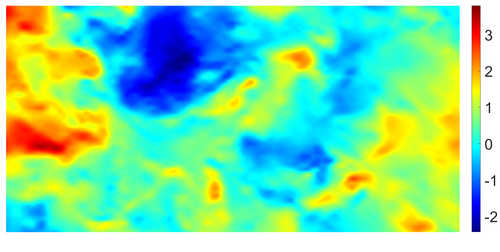}};
    \begin{scope}[x={(image.south east)},y={(image.north west)}]
        \draw[red, very thick] (0.55,0.85) rectangle (0.65,0.98);
        \draw[red, very thick] (0.02,0.08) rectangle (0.13,0.25);
    \end{scope}
    \end{tikzpicture}
    \begin{tikzpicture}
    \node[anchor=south west,inner sep=0] (image) at (0,0) {\includegraphics[height=3cm]{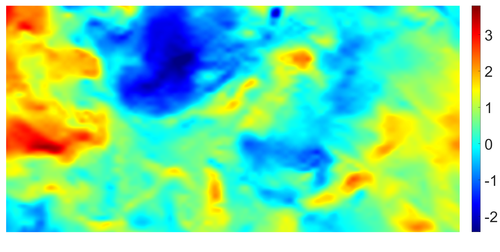}};
    \begin{scope}[x={(image.south east)},y={(image.north west)}]
        \draw[red, very thick] (0.52,0.85) rectangle (0.62,0.98);
        \draw[red, very thick] (0.02,0.08) rectangle (0.12,0.25);
    \end{scope}
    \end{tikzpicture} \\
 \vspace{-0.3cm}
  \includegraphics[trim={0 0 2cm 0},clip,height=3cm]{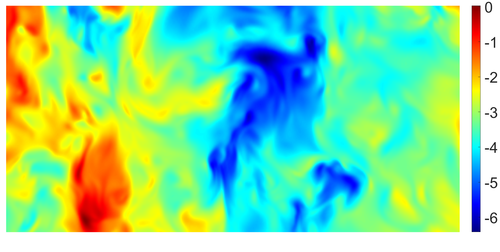}
 \includegraphics[trim={0 0 2cm 0},clip,height=3cm]{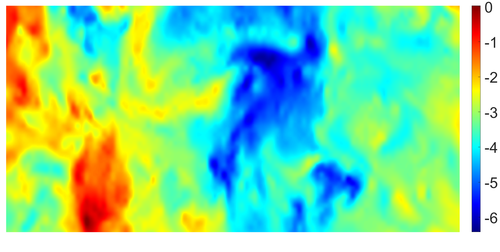}
 \includegraphics[height=3cm]{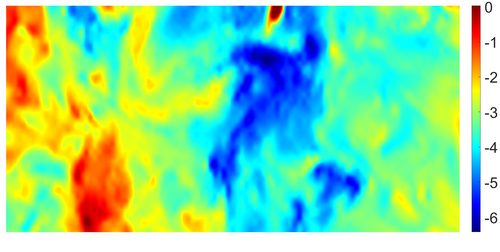} \\
 \vspace{-0.3cm}
   \includegraphics[trim={0 0 2cm 0},clip,height=3cm]{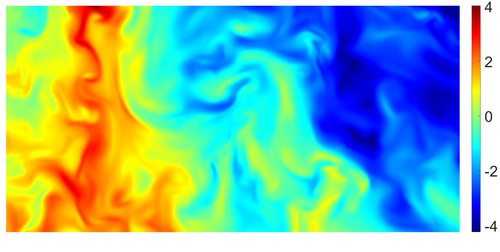}
 \includegraphics[trim={0 0 2cm 0},clip,height=3cm]{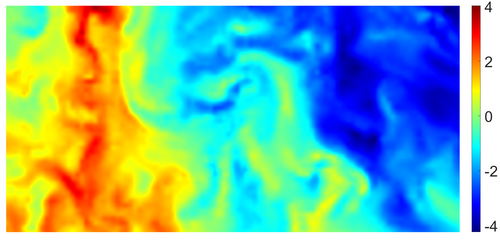}
 \includegraphics[height=3cm]{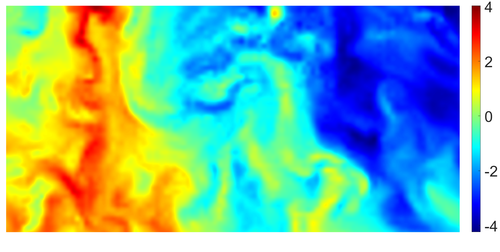} \\
 \vspace{-0.3cm}
   \includegraphics[trim={0 0 2cm 0},clip,height=3cm]{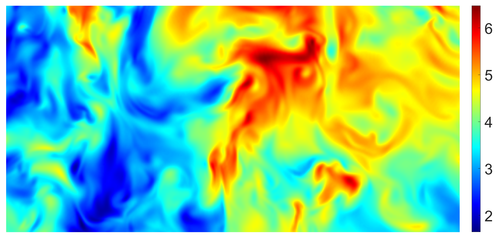}
 \includegraphics[trim={0 0 2cm 0},clip,height=3cm]{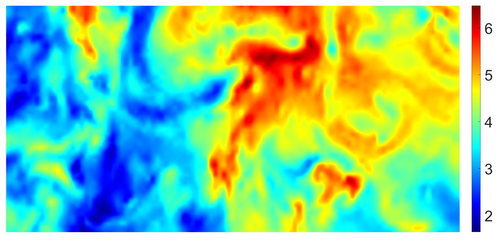}
 \includegraphics[height=3cm]{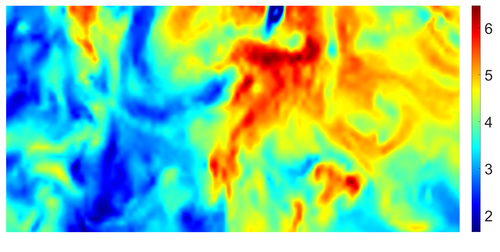} \\
 \includegraphics[trim={0 0 2cm 0},clip,height=3cm]{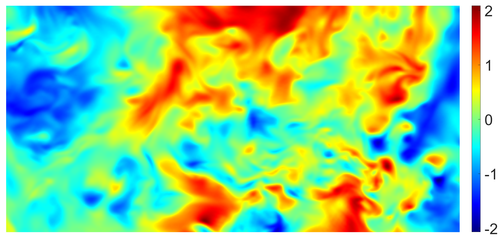}
 \includegraphics[trim={0 0 2cm 0},clip,height=3cm]{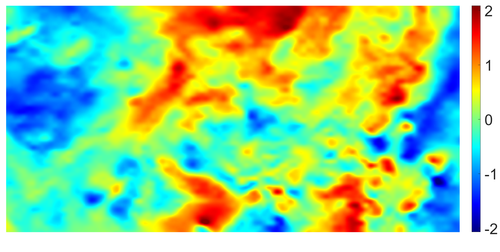}
 \includegraphics[height=3cm]{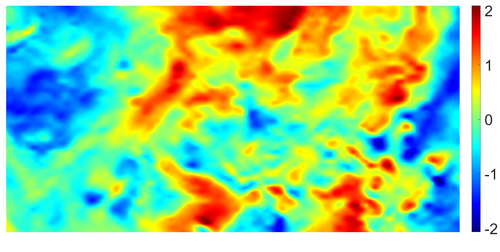} \\
 \vspace{-0.3cm}
  \includegraphics[trim={0 0 2cm 0},clip,height=3cm]{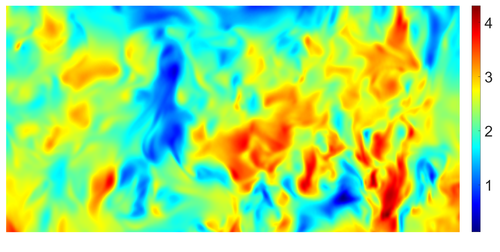}
 \includegraphics[trim={0 0 2cm 0},clip,height=3cm]{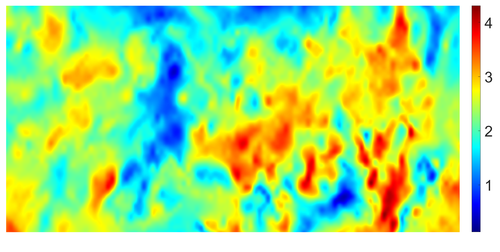}
 \includegraphics[height=3cm]{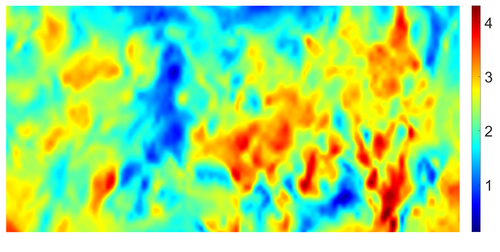} \\
 \vspace{-0.3cm}
   \includegraphics[trim={0 0 2cm 0},clip,height=3cm]{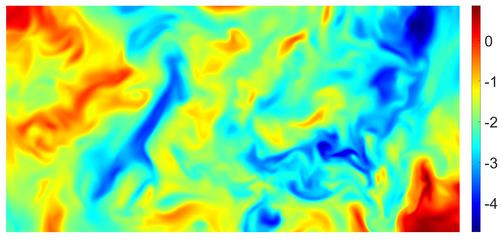}
 \includegraphics[trim={0 0 2cm 0},clip,height=3cm]{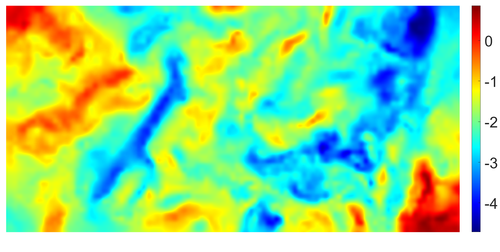}
 \includegraphics[height=3cm]{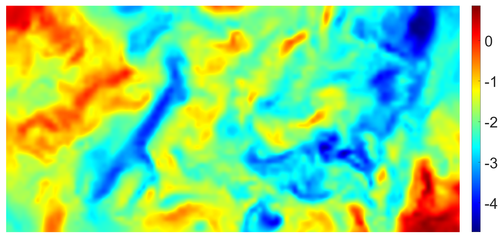} \\
 \vspace{-0.3cm}
   \includegraphics[trim={0 0 2cm 0},clip,height=3cm]{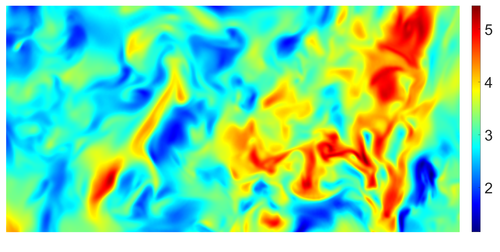}
 \includegraphics[trim={0 0 2cm 0},clip,height=3cm]{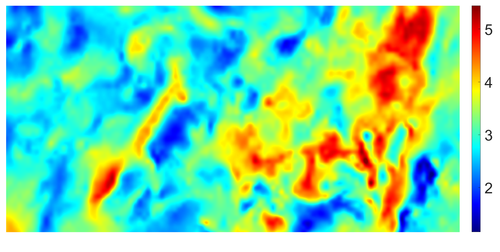}
 \includegraphics[height=3cm]{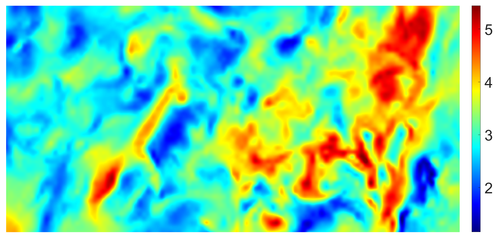} \\
    \vspace{-6mm}
 \caption{ Visualization of center $xy$-slice ($z=176$) of the flow for two of our tested datasets. \emph{Left to right:} Ground truth, $0.0125$ppp, $0.0225$ppp.
 \emph{Top to bottom:} X-,Y-,Z-component and magnitude.  
 Note that with higher particle density details are recovered better, but the method starts to fail in certain areas (see highlighted areas in first row).
 } \label{fig:results-ppp}
\end{figure*}

\paragraph{Particle Density \& Initialization Method.}
There is a trade-off for choosing the seeding density: A higher
density raises the observable spatial resolution, but at the same
time makes matching more ambiguous. This causes false positives,
commonly called ``ghost particles''.
Very high densities are challenging for all known reconstruction techniques.
The additive image formation model of \Eq \eqref{eq:ED} also
suggests an upper bound on the maximal allowed particle density.
\Tab\ref{tab:ppp} reports results for varying particle densities.
We measure recall (fraction of reconstructed ground truth particles)
and precision (fraction of reconstructed particles that coincide with
true particles to $<1$ pixel).
In \Fig\ref{fig:results-ppp} visualizations of our estimated flow fields for two different particle densities are shown.

To provide an upper bound, we initialize our method with ground truth
particle locations at time step 0 and optimize only for the flow estimation.
We also evaluate a sequential version of our method, in which we separate
energy \eqref{eq:Energy_functional} into particle reconstruction
and subsequent motion field estimation.
In addition to our proposed IPR-like triangulation, we
initialize particles with the popular volumetric tomography method
\textit{MART} \citep{els-06}. \textit{MART} creates a full, high-resolution
voxel grid of intensities (with, hopefully, lower intensities for
ghost particles and very low ones in empty space).
To extract a set of sub-voxel accurate 3D particle locations we perform
peak detection, similar to the 2D case for triangulation.
Since \textit{MART} always returns the same
particle set we run it only once, but increase the
number of iterations for the minimizer from $40$ to $160$.

Starting from a perfect particle reconstruction (\textit{true
particles}) the flow estimate improves with increasing particle
density.
Remarkably, our proposed iterative triangulation approach achieves
results comparable to the ground truth initialization, up to high
particle densities and is able to resolve most particle ambiguities.
In contrast, \textit{MART} and the sequential baseline struggle with
increasing particle density, which supports our claim that joint
energy minimization can better reconstruct the particles.

\paragraph{Sparsity Term.}
In \Tab\ref{tab:sparsity} we show a comparison between the two proposed sparsity norms and their influence on precision of the particle reconstruction and the flow endpoint error. The 0-norm performs slightly better than the 1-norm by eliminating more ghost particles that occur at high seeding densities.

\paragraph{Particle Size.}
For the above experiments, we have rendered the particles into the
images as Gaussian blobs with fixed $\sigma=1$, and the same is
done when re-rendering for the data term, respectively, proposal
generator.
We now test the influence of particle size on the reconstruction, by
varying $\sigma$. \Tab\ref{tab:partSize} shows results with hard
divergence constraint and fixed particle density $0.1$, for varying
$\sigma\in[0.6\hdots1.6]$. For small enough particles, size does not
matter, very large particles lead to more occlusions and
degrade the flow.
Furthermore, we verify the sensitivity of the method to unequal particle
size. To that end, we draw an individual $\sigma$ for each particle
from the normal distribution $\mathcal{N}(1,0.1^2)$, while still using
a fixed $\sigma=1$ during inference. As expected, the mismatch
between actual and rendered particles causes slightly larger errors.

\paragraph{Temporal Sampling.}
To quantify the stability of our method to different flow magnitudes
we modify the time interval between the two discrete time
steps and summarize the results in
\Tab\ref{tab:framerate}, together with the respective maximum flow
magnitude $|u|_2$. For lower frame rate (1.25x and 1.5x), and thus
larger magnitudes, we set our pyramid downsampling factor to $0.93$.

 \begin{figure*}[tb]
\begin{center}
 \includegraphics[width=0.48\textwidth]{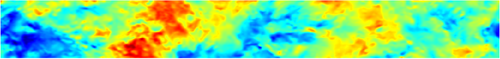}
 \includegraphics[width=0.48\textwidth]{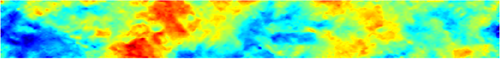}
 \includegraphics[width=0.025\textwidth]{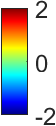}
  \\  
  \vspace{1mm}
  \includegraphics[width=0.48\textwidth]{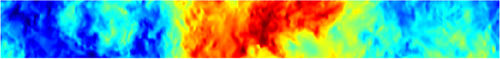}
  \includegraphics[width=0.48\textwidth]{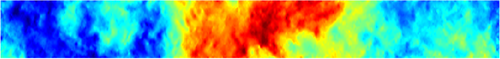}
  \includegraphics[width=0.025\textwidth]{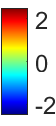}
    \\
  \vspace{1mm}
  \includegraphics[width=0.48\textwidth]{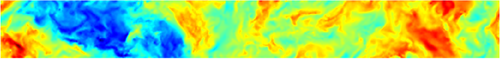}
  \includegraphics[width=0.48\textwidth]{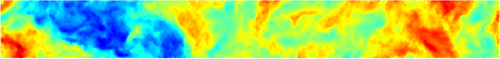}
  \includegraphics[width=0.025\textwidth]{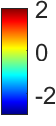}
  \end{center}
    \vspace{-4mm}
  \caption{ $xy$-slice of the flow field for snapshot 10 of
  the 4$^\text{th}$ PIV Challenge. \emph{Top to bottom:} $X,Y,Z$ flow
  components. \emph{Left:} multi-frame
  StB~\cite{sch-16}. \emph{Right:} our 2-frame method.}
  \label{fig:stb}
\end{figure*}

\begin{figure}[tb]
\begin{center}
  \includegraphics[width=\columnwidth]{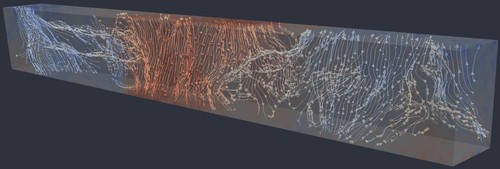} 
    \end{center}
    \vspace{-4mm}
   \caption{PIV Challenge: Streamline visualizations of our results for snapshot 10 of the 4$^\text{th}$ PIV Challenge~\cite{kah-16} with y component of the flow color coded.}
\label{fig:pivChall_streamline}
\end{figure}

\begin{table}[t]
\begin{center}
\caption{Influence of different sparsity norms: 0-norm vs. 1-norm ($0.2$ ppp).}
\label{tab:sparsity}
\vspace{\tablecaptionspace}
\begin{tabular}{ccccc}
\hline
\rowcolor{gray!10}
Norm & $|\cdot|_0$ & $|\cdot|_1$   \\ %
\hline
AEE & 0.108  & 0.110     \\
 \textit{prec.} & 55.43  &  25.23  \\
 \textit{recall} & 99.86 &  99.94 \\
\hline
\end{tabular}
\end{center}

\end{table}

\begin{table}[t]
\begin{center}
\caption{Influence of particle size on reconstruction quality (0.1 ppp).
}\label{tab:partSize}
\vspace{-\tablecaptionspace}
\setlength{\tabcolsep}{0.45em}
\begin{tabular}{cccccccc}
\hline
\rowcolor{gray!10}
{$\sigma$} & 0.6 & 0.8 & 1 & 1.2 & 1.4 & 1.6 & $\mathcal{N}_{1,0.1^2}$ \\
\hline
AEE &  0.194 &  0.135 & 0.136 & 0.140 & 0.217 & 0.235 & 0.155 \\
AAE & 3.388 & 2.465 &  2.486 & 2.561 & 4.002 & 4.575 & 2.879 \\
\hline
\end{tabular}
\end{center}
\end{table}

\begin{table}[t]
\begin{center}
\caption{Varying the sampling distance between frames ($0.1$ ppp).
\vspace{\tablecaptionspace}
}\label{tab:framerate}
\begin{tabular}{ccccc}
\hline
\rowcolor{gray!10}
Temp. dist. & 0.75x & 1.0x & 1.25x & 1.5x \\ %
\hline
AEE & 0.102 & 0.136 & 0.170 & 0.283  \\
Max. $|u|_2$ & 6.596 & 8.795 & 10.993 & 13.192 \\ %
\hline
\end{tabular}
\end{center}

\end{table}

\paragraph{PIV Challenge.}
Unfortunately, no ground truth is provided for the data of the
4$^\text{th}$ PIV Challenge~\cite{kah-16}, such that we cannot run a
quantitative evaluation on that dataset.
However, \citet{sch-16} kindly provided us results for
their method, StB, for snapshot 10. StB was the best-performing
method in the challenge with an endpoint error of $\approx$0.24 voxels
(compared to errors $>$0.3 for all competitors). The average endpoint
difference between our approach and StB is $<$0.14 voxels.
In \Fig\ref{fig:stb} both results appear to be visually comparable, yet,
note that StB includes a tracking procedure that requires data of multiple
time steps ($15$ for the given particle density $0.1$). %
\Fig\ref{fig:pivChall_streamline} shows a streamline visualization of our results at snapshot 10.

\section{Experimental Data}
\label{sec:expData}

We show qualitative results of two experiments in both air and water, utilizing the pinhole camera model and the polynomial camera model respectively.

\begin{figure}[tb]
\begin{center}
  \includegraphics[width=\columnwidth]{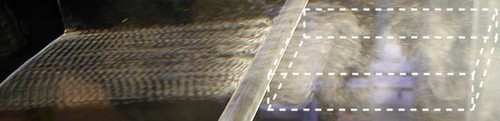} \\
  \vspace{1mm}
  \includegraphics[width=\columnwidth]{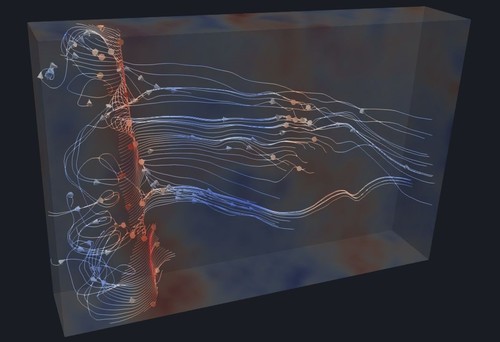} \\
  \vspace{1mm}
    \includegraphics[width=\columnwidth]{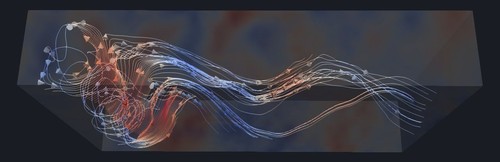} 
    \end{center}
    \vspace{-4mm}
   \caption{Experimental setup in air: \emph{Top:} Karman street behind cylinder. \emph{Middle and bottom:} Streamline visualizations of our results with color coding based on z component of the flow (where the Karman street can be observed best).}
\label{fig:expData_streamline}
\end{figure}

\begin{figure}[tb]
\begin{center}
  \includegraphics[width=0.49\columnwidth]{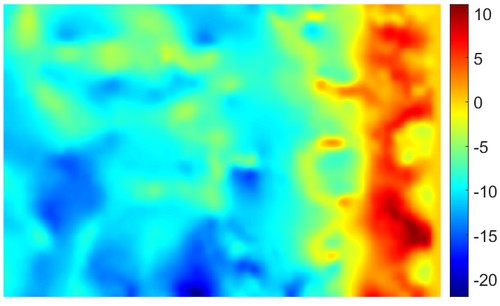} 
    \includegraphics[width=0.49\columnwidth]{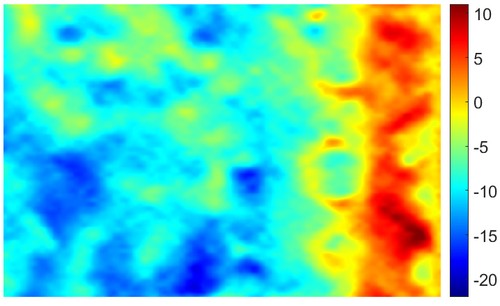} 
 \\
    \includegraphics[width=0.49\columnwidth]{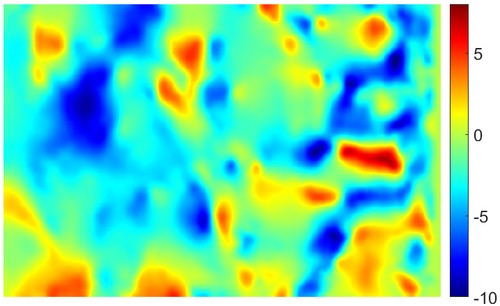} 
    \includegraphics[width=0.49\columnwidth]{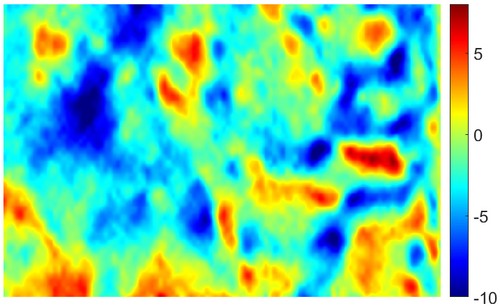} 
     \\  
      \includegraphics[width=0.49\columnwidth]{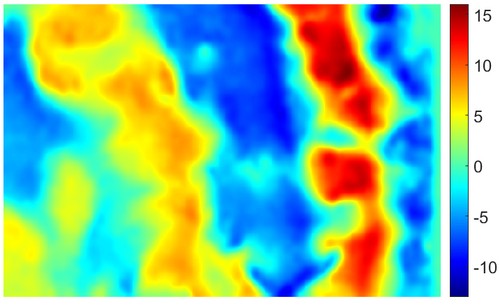} 
      \includegraphics[width=0.49\columnwidth]{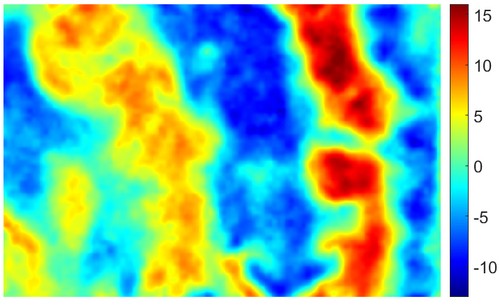} 
\\

\end{center}
    \vspace{-4mm}
   \caption{Experimental setup in air: $xy$-slice of the flow at $z=203$. \textit{Left:} Reference flow field provided by LaVision (TomoPIV). \textit{Right:} Results with our approach. (\textit{top} to \textit{bottom}: flow in $X$, $Y$ and $Z$-direction).}
\label{fig:expData}
\end{figure}

\paragraph{Experiment in air.}
We test our approach on experimental data provided by \emph{LaVision}\footnote{Package 9 in \url{http://fluid.irisa.fr/data-eng.htm}} (see \cite{mic-06}). The experiment captures the wake flow behind a cylinder, which forms a so-called Karman vortex street (see \Fig\ref{fig:overview} and \ref{fig:expData_streamline}). Data is provided in form of a tomographic reconstruction of the particle volume. In order to run our method, we backproject the particle volume to four arbitrary camera views (we take the same as for our simulated dataset) and use those backprojected images as input to our method. 
Since particle densities allow it and the provided reference flow is of low resolution, we downsample the input volume by a factor of 2 from $2107 \times 1434 \times 406$ to $1054 \times 717 \times 203$ and render to 2D images of dimensions $1500 \times 1100$ with particle size $\sigma = 0.6$.
Note, that since those camera locations do not necessarily coincide with the original camera locations, ghost particles in the reconstructed volume may lead to wrong particles in the backprojected image. %
However, as results in \Fig\ref{fig:expData} indicate, our algorithm is able to recover a detailed flow field that corresponds with the reference flow despite these deflections in the data. In addition to our own result, we show results provided by LaVision. The provided reference flow field was estimated with TomoPIV \citep{els-06}, using a final interrogation volume size of $48^3$ and an overlap of $75\%$, \ie one flow vector at every 12 voxel locations. It can be seen in \Fig\ref{fig:expData} that our method recovers much finer details of the flow, due to the avoidance of large interrogation volumes.
In order to cope with flow magnitudes up to 15.5 voxel we chose 16 pyramid levels and a pyramid downsampling factor of $0.92$.
Note, that in the resulting flow field the cylinder is positioned to the right of the volume and the general flow direction is towards the left. Effects of the Karman vortex street can be primarily seen in the z component of the flow (periodically alternating flow directions with decreasing magnitude from right to left).
\begin{figure}[t!]
\begin{center}

  \includegraphics[width=\columnwidth]{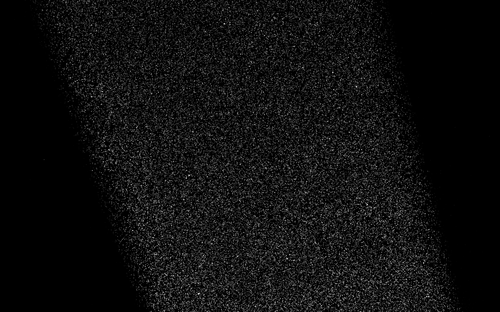} \\
  \vspace{1mm}
  \includegraphics[width=\columnwidth]{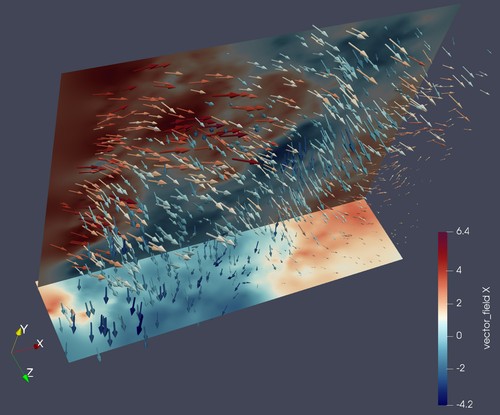} \\
  \vspace{1mm}
  \includegraphics[width=\columnwidth]{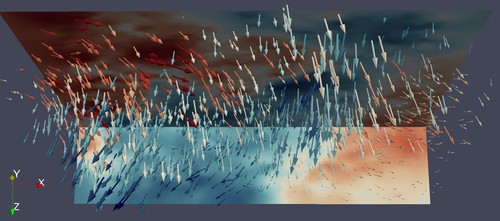}
\end{center}
    \vspace{-4mm}
   \caption{Experimental setup in water: \emph{Top:} Input camera image for one view. \emph{Bottom:} Visualization of flow vectors with X-flow for color coding and flow magnitude for vector size.}
\label{fig:schanz_input}
\end{figure}

\paragraph{Experiment in water.}
To test our polynomial camera model we show results of an isotropic turbulence experiment in water.
The experimental setup includes a cylinder filled with water and two discs, located on the top and bottom of the cylinder, which are rotating in opposite direction. This setup is also known as \emph{French washing machine}. The rotating discs lead to a turbulent flow, with multiple rotational patterns.
In \Fig\ref{fig:schanz_input} we show one of the input camera images together with visualizations of the resulting flow field, obtained from two time steps.
The arrow size encodes the magnitude of the flow and the color denotes its motion in X-direction (red positive, blue negative). One can see a clock-wise rotating vortex on the left side of the volume and a counter-clockwise rotating vortex with lower magnitude on the right.
The camera was calibrated from 20000 point correspondences, which were kindly provided together with the data by Daniel Schanz (DLR). Following \citet{har-97}, we use DLT to get an initial estimate of the 38 parameters of the polynomial camera model and optimize the result using the iterative Levenberg-Marquardt method.
Input images have a resolution of $2560 \times 2600$.
Based on the given point correspondences and the image resolution we chose a down-sampled volume of $162 \times 114 \times 36$ voxels for our motion field estimation.

\section{Conclusion}
\label{sec:conclusion}

We have presented the first variational model that \emph{jointly}
solves sparse particle reconstruction and dense 3D fluid motion
estimation in PIV/PTV data for the common multi-camera setup.
The sparse particle representation allows us to utilize the
high-resolution image data for matching, while keeping memory
consumption low enough to process large volumes.
Densely modeling the fluid motion in 3D enables the direct use of
physically motivated regularizers, in our case viscosity and incompressibility.
The proposed joint optimization 
captures the mutual dependencies between
particle reconstruction and flow estimation. This yields results that
are clearly superior to traditional, sequential
methods \citep{els-06,las-17}; and, using only 2 frames, competes with
the best available multi-frame methods, which require sequences of
15-30 time steps.
We have validated the performance of our approach both quantitatively on synthetic data
and qualitatively on real experiments in water and air.

In our experiments we have demonstrated that the proposed joint formulation allows for much 
higher seeding densities than traditional sequential approaches, by implicitly utilizing information of 
both time steps also for the particle reconstruction, thus reducing the amount of wrongly reconstructed ghost particles.

\changes{
One limitation of our current regularization scheme is that it does
not account for turbulent (non-Stokes) flow of high Reynolds numbers.
Here, an extension in the spirit of \citet{ruh-07} could be a
promising future direction.  Another interesting extension is to put
more focus also on the recovery of the pressure field. To that end one
might again follow \citet{ruh-07} and move to a discretization scheme
with mixed finite elements, which fulfills the Babuska-Brezzi
condition~\citep{bre-12}.
}
Our approach could further be extended to more than two time steps to even better exploit temporal consistency.
Besides resolving some of the remaining ambiguities of the particle reconstruction, this would also facilitate the use of additional physical constraints, \eg, incorporating the transport equations (inertial part) of the Navier-Stokes model \citep[\cf][]{ruh-06}.
Such an integrated model over multiple time steps will considerably increase the computational cost, and may require additional efforts to make energy minimization more efficient.
Also, when dealing with longer sequences one will have to account for the possibility that 
particles enter or leave the measurement volume, and for particles' intensity changes over time.

\paragraph{Acknowledgements.}
This work was supported by ETH grant 29~14-1. Thomas Pock and Christoph Vogel
acknowledges support from the ERC starting grant 640156, ’HOMOVIS’.
We thank Daniel Schanz for kindly sharing their results on the 4th PIV Challenge and for providing experimental data in water.

\clearpage

\bibliographystyle{apalike}      %
\bibliography{references}   %

\end{document}